\documentclass{IEEEtaes}

\usepackage{color,array,amsthm}
\usepackage{graphicx}
\usepackage{amssymb} % The amssymb package provides various useful mathematical symbols
\usepackage{amsmath} % The amsmath package provides various useful equation environments.
\let\labelindent\relax
\usepackage{enumitem}
\usepackage{hhline}
\usepackage{bm}
\usepackage{xcolor}
\usepackage{url}

\jvol{XX}
\jnum{XX}
\jmonth{XXXXX}
\paper{1234567}
\pubyear{2025}
\doiinfo{TAES.2025.Doi Number}

\begin{document}

\title{POLAR-Sim: Augmenting NASA's POLAR Dataset for Data-Driven Lunar Perception and Rover Simulation} 

\author{Bo-Hsun Chen}
\affil{}

\author{Peter Negrut*}
\affil{}

\author{Thomas Liang*}
\affil{}

\author{Nevindu Batagoda}
\affil{}

\author{Harry Zhang}
\affil{}

\author{Dan Negrut}
\affil{}

\receiveddate{Manuscript received XXXXX 00, 0000; revised XXXXX 00, 0000; accepted XXXXX 00, 0000. This work was supported in part under NASA project STTR 80NSSC24CA030.\\
%% This paragraph of the first footnote will contain the date on which you submitted your paper for review, which is populated by IEEE. It is IEEE style to display support information, including sponsor and financial support acknowledgment, here and not in an acknowledgment section at the end of the article. For example, ``This work was supported in part by the U.S. Department of Commerce under Grant BS123456.'' 
}
%% \accepteddate{XXXXX XX XXXX}
%% \publisheddate{XXXXX XX XXXX}

%% \corresp{The name of the corresponding author appears after the financial information, e.g. {\itshape (Corresponding author: M. Smith)}. Here you may also indicate if authors contributed equally or if there are co-first authors.}
\corresp{*: equal contribution.}

\authoraddress{
All authors are with the Simulation-Based Engineering Lab,
University of Wisconsin-Madison, Madison, WI-63706, USA
{\tt\small \{bchen293, pnegrut,thomas.liang, batagoda, hzhang699, negrut\}@wisc.edu}
%% The next few paragraphs should contain the authors' current affiliations, including current address and e-mail. For example, First A. Author is with the National Institute of Standards and Technology, Boulder, CO 80305 USA (e-mail: \href{mailto:author@boulder.nist.gov}{author@boulder.nist.gov}). Second B. Author, Jr., was with Rice University, Houston, TX 77005 USA. He is now with the Department of Physics, Colorado State University, Fort Collins, CO 80523 USA (e-mail: \href{mailto:author@lamar.colostate.edu}{author@lamar.colostate.edu}). Third C. Author is with the Electrical Engineering Department, University of Colorado, Boulder, CO 80309 USA, on leave from the National Research Institute for Metals, Tsukuba 305-0047, Japan (e-mail: \href{mailto:author@nrim.go.jp}{author@nrim.go.jp}).
}

\editor{Supplemental video for demonstration is included.}
\supplementary{Color versions of one or more of the figures in this article are available online at \href{http://ieeexplore.ieee.org}{http://ieeexplore.ieee.org}.}

\markboth{AUTHOR ET AL.}{SHORT ARTICLE TITLE}
\maketitle

\begin{abstract}
NASA's POLAR (Polar Optical Lunar Analog Reconstruction) dataset contains approximately 2,600 pairs of high dynamic range stereo photos captured across 13 varied terrain scenarios, including areas with sparse or dense rock distributions, craters, and rocks of different sizes. The purpose of these images is to spur research and development in robotics, AI-based perception, and autonomous navigation. Acknowledging a scarcity of lunar images from around the lunar poles, NASA Ames produced on Earth but in controlled conditions images that resemble rover operating conditions from these regions of the Moon. This contribution reports outcomes of two tasks. In Task 1, we provided bounding boxes and semantic segmentation information for all the images in NASA's POLAR dataset. This effort resulted in 23,000 labels and semantic segmentation information pertaining to rocks, shadows, and craters. In Task 2, we generated the digital twins of the 13 scenarios that have been used to produce all the images in the POLAR dataset. Specifically, for each of these scenarios, we produced individual meshes, texture information, and material properties associated with the ground and the rocks in each scenario. This allows anyone with a camera model to generate synthetic images associated with any of the 13 scenarios of the POLAR dataset. Effectively, one can generate as many semantically labeled synthetic images as desired -- from different locations in the scene, with different exposure values, for different positions of the sun, with or without the presence of active illumination, etc. The benefit of this work is twofold. Using outcomes of Task 1, one can train and/or test perception algorithms that deal with Moon images. For Task 2, one can produce as much data as desired to train and test AI algorithms that are anticipated to be used in lunar conditions. All the outcomes of this work are available in a public repository for unfettered use and distribution.
\end{abstract}

%% \begin{IEEEkeywords}
%% \end{IEEEkeywords}
\begin{IEEEkeywords}
%% Enter keywords or phrases in alphabetical order, separated by commas. For a list of suggested keywords, send a blank e-mail to \href{mailto:keywords@ieee.org}{keywords@ieee.org} or visit \href{http://www.ieee.org/organizations/pubs/ani\_prod/keywrd98.txt}{\url{http://www.ieee.org/organizations/pubs/ani\_prod/keywrd98.txt}}
	lunar ground exploration, object detection, perception, semantic segmentation, synthetic images, training ground truth data 
\end{IEEEkeywords}

\section{INTRODUCTION}
\label{sec:intro}
\subsection{Motivation}
\label{subsec:motivation}
% Bo-Hsun to take first shot. Talk here about POLAR dataset: why it exists, and what it contains. About 0.5 pages.

Renewed interest in lunar exploration is on the rise, driven by the Moon's potential as a staging ground for future missions to more distant celestial bodies, such as asteroids or Mars. The pace of Moon-focused missions is accelerating, with countries like the United States, China, India, Russia, and Japan either having successfully completed lunar landings or planning upcoming missions. Simulation plays a critical role in the success of these missions, as testing under true lunar conditions is impractical. The lunar environment presents unique challenges, including lower gravity, an almost complete lack of atmosphere, distinct lighting conditions, different terramechanics compared to Earth, and the unique composition of the lunar surface. These factors are difficult to replicate on Earth, necessitating a reliance on simulation environments \cite{pangu2004,brochard2018scientific,allan2019planetary,muller2021photorealistic,sewtz2022ursim,pieczynski-lunarsim2023}. Simulation environments allow for the easy generation and testing of perception, planning, and control solutions, offering advantages in reducing costs and time to solution.

Our contribution is motivated by the observation that camera-based perception tasks on the Moon differ significantly from those on Earth. The lunar surface is covered with lunar regolith, a low-albedo and retroreflective material that influences the reflectivity of the surface. The absence of atmospheric scattering and the properties of the lunar regolith are key factors shaping lunar lighting conditions, which result in hard lighting, long shadows, low sunlight angles, and high dynamic range (HDR) contrasts between illuminated and shadowed regions \cite{hapke1981bidirectional, hapke2008bidirectional, sato2014resolved}. A particularly notable phenomenon is the \textit{opposition effect}, where there is a marked increase in brightness when observing the lunar surface from a direction aligned with an illumination source, most often the Sun. The primary cause of the opposition effect is shadow hiding, where shadows disappear when the viewing and illumination directions are the same \cite{kuzminykh2021physically}. %While several papers have proposed integrated simulation platforms for extraterrestrial exploration by incorporating high-fidelity virtual cameras and dynamic rover simulation on the Moon \cite{allan2019planetary} and Mars \cite{sewtz2022ursim}, these studies have rarely employed data-driven approaches to construct digital scenarios or verify simulations with GT approval.

To encourage the development of vision perception algorithms for lunar environments, NASA released the \textit{Polar Optical Lunar Analog Reconstruction (POLAR)} dataset \cite{wong2017polar}, which consists of Earth-based photos that replicate the visual perception conditions of polar lunar lighting. The POLAR dataset contains approximately 2,600 pairs of HDR stereo photos and LiDAR-scanned point cloud data for 13 terrain scenarios, capturing typical lunar landscapes featuring pits, mounds, rocks, fresh craters, regolith, and rock and regolith materials. The dataset was intended for developing and validating stereo camera algorithms for lunar applications. 

%\textcolor{blue}{Although the existing digital elevation models (DEMs) of the lunar surface, which were scanned by the Lunar Orbiter Laser Altimeter (LOLA) \cite{BARKER2016346}, provide detailed geometry, they lack real photos for GT rendering validation. That is, although in typical vehicle simulation for lunar exploration, one can run the rover on the DEMs using the Hapke BRDF for virtual camera rendering, it lacks a GT approval for evaluating the quality of the rendered images. In contrast, the POLAR dataset includes real photos for rendering validation, making it valuable for simulation-based evaluations. Additionally, the POLAR dataset specifies the number of rocks in each terrain scenario, aiding annotators in accurately locating and separating rocks from the ground. When treating the rocks as obstacles, such rock separation is crucial for training data-driven hazard detection algorithms and for constructing simulation scenarios with assigning distinct visual and dynamic characteristics for meshes of rocks and ground. Thus, our POLAR-Sim dataset is better suited for integrating camera and vehicle dynamics simulations than using the actual lunar DEMs alone.}

\subsection{Contribution}
\label{subsec:contribution}
% Bo-Hsun to take first shot. About 0.25 pages. Explain here what we have accomplished. This should mirror what's in the Abstract. Say that somebody went into a studio and created these photographs. Basically, we regenerated the world that was produced in the studio.

Our work augments NASA's POLAR dataset with ground-truth labels and digital assets to assist computer vision researchers in enhancing lunar perception algorithms anchored in data-driven methods. For all the images in the POLAR dataset, we generated digital assets that facilitate the training and evaluation of machine learning algorithms for lunar perception tasks. In a second task, in which we generated a digital twin of each POLAR scenario, we separated the LiDAR point clouds of the ground and rocks and generated geometric meshes for all assets in the scenarios. We manually annotated bounding boxes and semantic segmentation maps for the ground, as well as for approximately 23,000 rocks and their shadows, across all 2,600 pairs of POLAR HDR stereo photos, providing ground truth labels for machine learning-based perception algorithms. As a demonstration of the utility of our labeling and digitization efforts, we: (\textit{i}) conducted both camera and ground vehicle dynamics simulations in the POLAR digital twin scenarios, and (\textit{ii}) used the rock and shadow labels to train visual perception algorithms under lunar lighting conditions. Our contributions are as follows:
\begin{itemize}[leftmargin=*]
	\item We introduce \textit{POLAR-Sim}, a publicly available dataset containing bounding box and semantic segmentation labels (in YOLO format) for the ground, rocks, and rock shadows, as well as meshes for all separated ground and rock elements of the 13 terrain scenarios in the POLAR dataset. POLAR-Sim is publicly accessible on GitHub \cite{polarsimChen2023}.
	
	%\item We propose a new metric called the \textit{Instance Performance Difference (IPD)} to measure the simulation-to-reality (sim-to-real) gap. This metric more accurately quantifies the gap by assessing the task performance difference between the real and synthetic datasets from the perspective of a perception algorithm.
	
	\item We present a data-driven pipeline for digitizing the POLAR dataset. This pipeline enables the creation of digital twins for all 13 POLAR scenarios. These digital twins facilitate the synthesis of photorealistic images via camera simulation and the testing of vision perception algorithms. We validate the photorealism of the synthetic images and the fidelity of the digitized terrain by means of YOLOv5 \cite{jocher_yolov5_2020} and the instance performance difference (IPD) metric \cite{chen2024instance}.
	
	\item We demonstrate the use of the digitized terrains by ground vehicles, which operate in POLAR terrain conditions using an in-house developed dynamics simulation engine. Our demonstration video highlights the Volatiles Investigating Polar Exploration Rover (VIPER) \cite{overviewVIPER2019}, specifically designed for operation near the lunar south pole, as it traverses in various simulated POLAR terrain scenarios and maneuvers over rocks. The front-end virtual camera produces the synthetic images used by YOLOv5 for hazard detection operations, demonstrating the POLAR-Sim facilitated integration of terramechanics and computer vision.
\end{itemize}

% To our best knowledge, this is the first effort to label photos and digitize terrains within the POLAR dataset for broader robotic research applications. Additionally, this is a pioneering paper to present a high-level data-driven simulation of ground vehicle operations that integrates a high-fidelity terramechanics dynamics model with photorealistic image synthesis for visual perception.

% This paper is an extended version of parts of our conference paper \cite{batagoda2024physics-based}. While the object detection task has been addressed in the conference paper, only three terrain scenarios (Terrains 1, 4, and 11) were included as a pilot study. This paper presents results from all the terrain scenarios in the POLAR dataset with a more comprehensive discussion to complete the study.

The paper is organized as follows. Section \ref{sec:polarsim_components} details the labeling methods and the process used to create digital twins. Section \ref{sec:use_cases} introduces three use cases of POLAR-Sim: training a visual object detection algorithm, synthesizing and verifying photorealistic images, and demonstrating the VIPER rover's traversal of virtual POLAR terrains. The results obtained are discussed in Section \ref{sec:result_discussion}, which also briefly surveys the state-of-the-art. Conclusions and directions of future work are provided in Section \ref{sec:conclusion}.

\section{POLAR-Sim COMPONENTS}
\label{sec:polarsim_components}
The POLAR-Sim dataset builds off NASA's POLAR dataset, which was developed using a lunar analog terrain in a sandbox populated with rocks, pits, and hills \cite{wong2017polar}. The ground and rocks were covered with fine and powdery sand called regolith simulant, and a tungsten-halogen spotlight simulated the Sun. This setup replicated the low-elevation sunlight at the lunar poles, the rugged terrain, and the reflectance properties of the lunar regolith surface. POLAR-Sim consists of three components: photo bounding box annotations, photo semantic segmentation annotations, and meshes of the ground and rocks.

\subsection{Photo Bounding Box Annotation}
\label{subsec:photo_bbox_labels}
%\SBELcomment{Bo-Hsun} please use the NASA tech report to dump stuff in here. 
%Explain here the labeling process -- what we labeled and how it can be used. Show a couple of pics. In total, should be one page at most.

To support the training of data-driven perception algorithms, we manually labeled bounding boxes for all of the rocks and their shadows in the POLAR dataset. This effort was motivated by the observation that object detection for rocks and shadows plays an important role in autonomous navigation -- large rocks can block the rover's path, while medium and small rocks can damage the wheels or the chassis. Shadows also help estimate the Sun's position, which is vital for navigation planning, solar energy harvesting, and sensor orientation.

Approximately 23,000 rocks and rocks' shadows were labeled. Each photo's configuration includes the terrain ID, stereo camera position (A: 1.5 m from terrain center at 0$^\circ$, B: 4 m from terrain center at 0$^\circ$, or C: 1.5 m from terrain center at 280$^\circ$), rover light status (ON or OFF), Sun azimuth angle (none, 30, 180, 270, or 350 degrees), stereo camera index (Left or Right), and exposure time (32 to 2048 ms), where ``none'' means no simulated Sun used. Each POLAR photo was taken under a combination of these configuration parameters. Note that the labels for the rocks and shadows remained the same for several POLAR photos. Specifically, exposure time variations did not alter the positions of the rocks and their shadows, so photos with exposure times of 32, 64, and 128 ms share the same labels. Other similarities include: different rover light statuses have the same rock and shadow labels, same camera positions with different Sun azimuths have the same rock labels but different shadow labels, and the Left and Right camera views have similar rock and shadow label positions. Leveraging these observations reduced the burden of labeling. Finally, since in the POLAR dataset the exposure time was controlled, at very low exposure time the shadow labels were deleted subjectively by the human annotator, since it was dark enough for the shadow to be judged as ``invisible.''

\begin{figure}[!t]
	\centering
	\includegraphics[width=1.0\columnwidth]{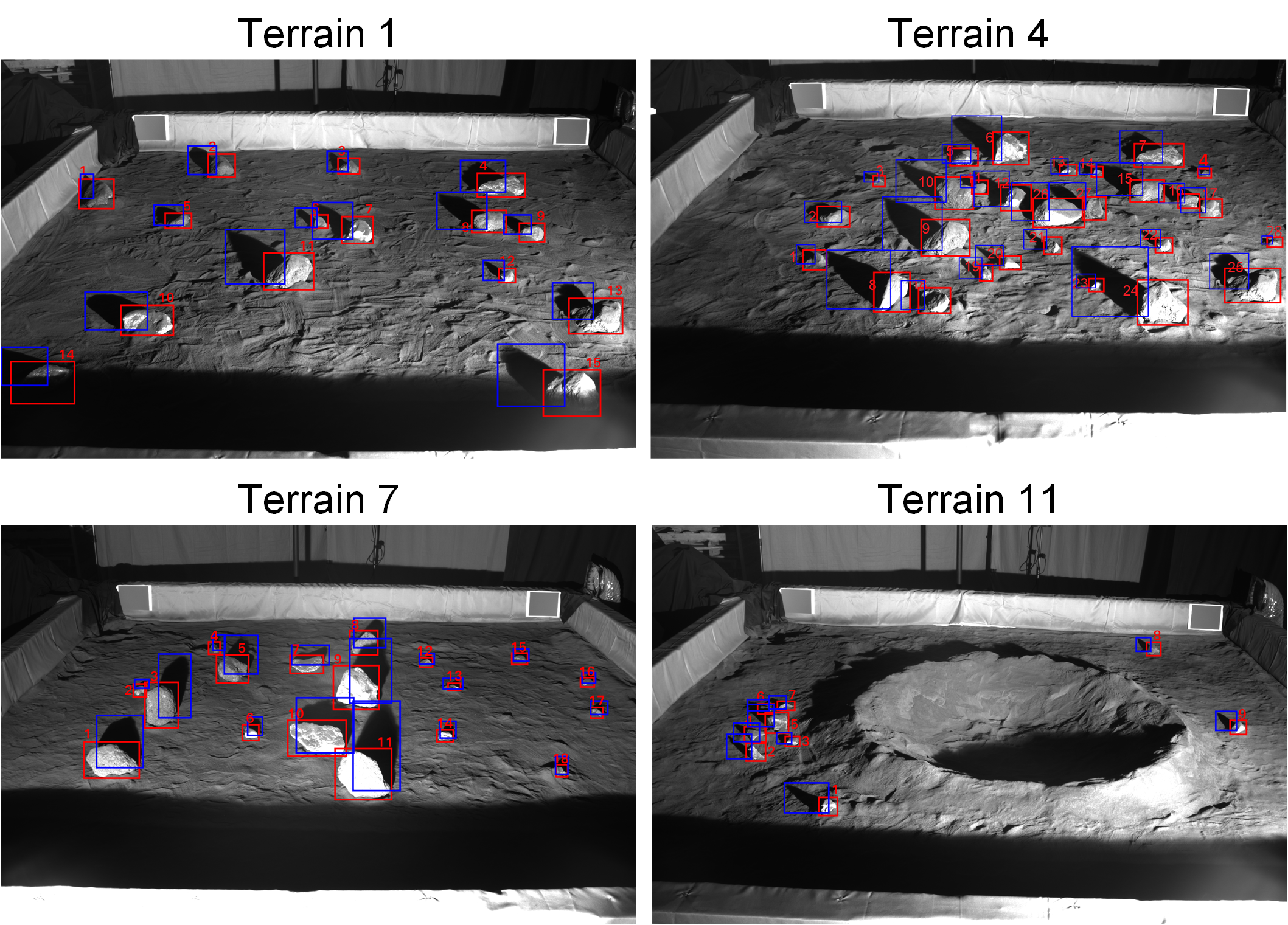}
	\caption{Illustrations of rock (red) and shadow (blue) bounding box annotations.}
	\label{fig:bbox_illu}
\end{figure}

Annotation was performed using \textit{labelImg} \cite{tzutalin2015labelimg}. For each terrain, the first Left stereo photo with rover light OFF was manually labeled. Labels were reused for photos with different exposure times, with shadow labels omitted for very low exposures due to their invisibility. The labels were then replicated and adjusted for the Right camera view with similar rock and shadow positions, and were replicated for photos with different Sun azimuths which only required shadow adjustments. Thereafter, labels were replicated and fine-tuned for photos with rover light ON. This process was repeated for all stereo camera positions (A, B, and C) and for all 13 POLAR terrain scenarios. The annotations were saved in YOLO format as TXT files. Figure~\ref{fig:bbox_illu} illustrates several examples of bounding box annotations.

\subsection{Photo Semantic Segmentation Map Annotation}
\label{subsec:photo_semantic_labels}
\begin{figure}[!t]
	\centering
	\includegraphics[width=1.0\columnwidth]{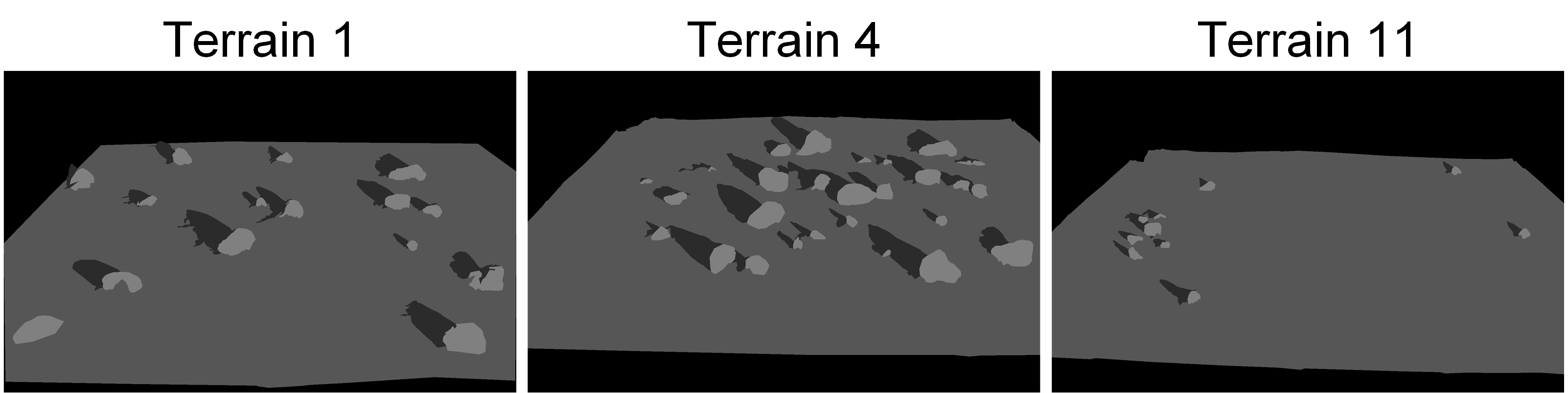}
	\caption{Illustrations of semantic segmentation map annotations. Colors from the lightest to darkest mark the rocks (128), ground (86), rock shadows (43), and background (0), respectively, where the numbers in the parentheses are the gray-scale values in 8-bit-depth.}
	\label{fig:semantic_illu}
\end{figure}

In POLAR-Sim, we provided semantic segmentation labels, manually annotating the background, ground, rocks, and rock shadows in the photos using, to the extent possible, Roboflow \cite{dwyer2024roboflow}. A similar procedure to the bounding box annotation was applied, leveraging repeated segmentation across photos of similar configurations. For each terrain, the clearest photo with suitable brightness was selected for accurate annotation, refined after using the Segment Anything Model (SAM) \cite{segmentAnything2023}. The resulting segmentation files were then copied and fine-tuned for other photos of similar configurations. This method ensured consistent and high-quality segmentation across the dataset. Segmentation annotations were also saved in YOLO format as TXT files, with a converter program provided to output the conventional gray-scale map formats. Figure~\ref{fig:semantic_illu}  shows examples of the semantic segmentation maps.

\subsection{Mesh Construction of the Ground and Rocks}
\label{subsec:mesh_construct}
% Explain here what we have done to generate virtual worlds. Show a couple of pics. Show a couple of pics. In total, should be one page at most. \SBELcomment{Bo-Hsun} please use the NASA tech report to dump stuff in here.

In the Task 2 of this effort, we produced mesh files of the ground and rocks for each terrain scenario in the POLAR dataset. Manually locating the rocks and generating surface meshes was performed in MATLAB, using the point cloud data provided in the dataset \cite{wong2017polar}. First, for each terrain scenario, two point clouds, respectively scanned from camera positions A and C (positions were defined in the 2nd paragraph in Sec.~\ref{subsec:photo_bbox_labels}), were inversely transformed back to the sandbox coordinates (where +X corresponds to Sun azimuth 0 deg, +Y to 90 deg, and +Z is upward). To enhance the mesh completeness, e.g., mitigating voids and occlusions from a single viewpoint, and to better capture the 3D shape of each rock, the two point clouds were manually coarse-aligned and positioned at the origin. Each rock's point clouds were then further locally fine-aligned to minimize occlusions and recover as much of the rock's shape as possible. Finally, the annotators manually identified the X, Y, and Z coordinate ranges of the rock to form a bounding cuboid.

\begin{figure}[!t]
	\centering
	\includegraphics[width=1.0\columnwidth]{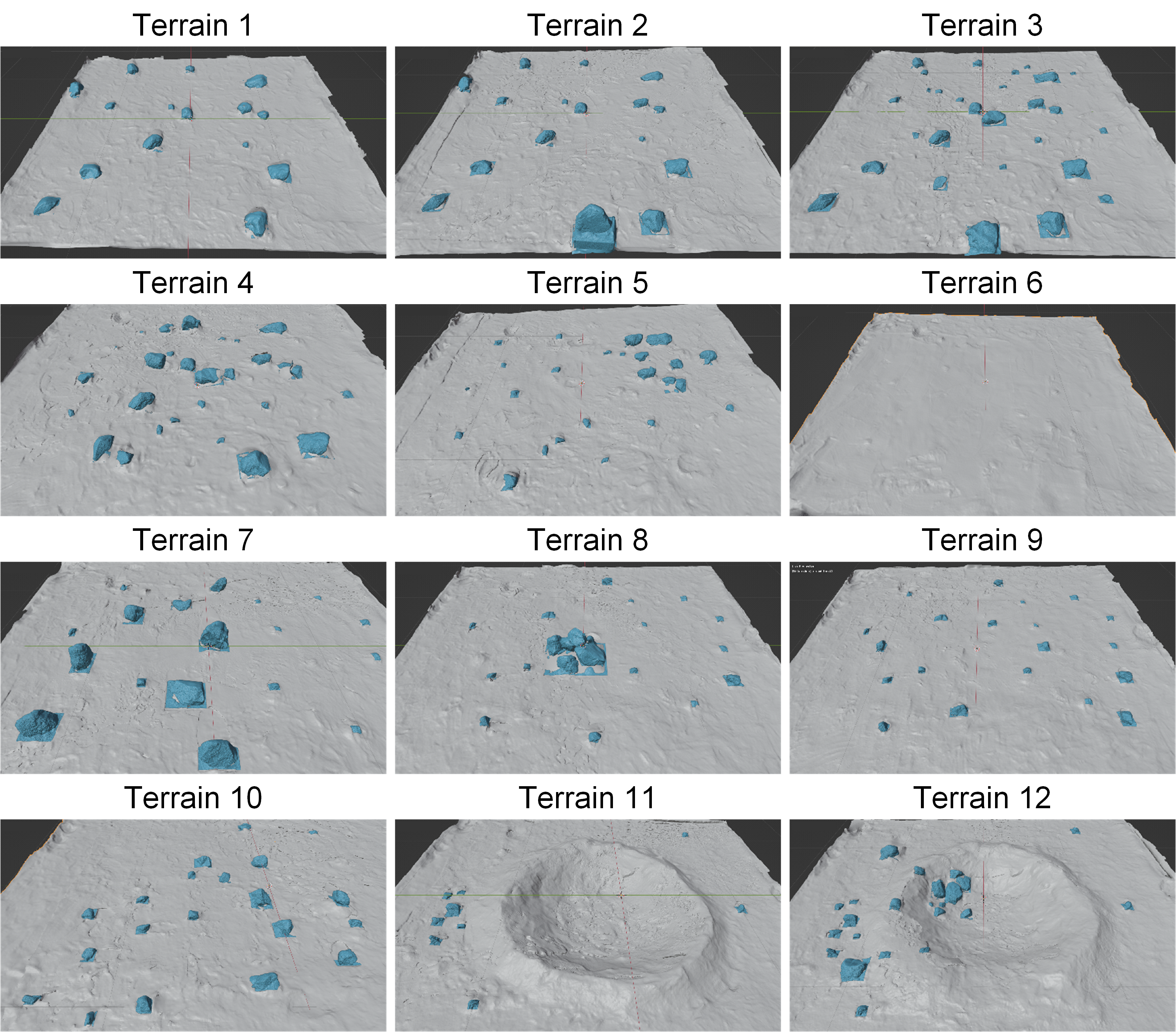}
	\caption{Meshes of the ground and rocks of 12 of the 13 POLAR-Sim scenarios. Blue color marks the rocks, and the gray color marks the ground. Each rock has its own mesh.}
	\label{fig:mesh_demo}
\end{figure}

Once all the rocks in the terrain were located, the point clouds of the rocks and ground were separated. These separated point clouds were then converted into meshes using the Poisson Surface Reconstruction method in MATLAB \cite{kazhdan2006poisson}, and the results were stored as OBJ files. Figure~\ref{fig:mesh_demo} illustrates 12 of the 13 POLAR-Sim terrain meshes.

\section{SAMPLE USE CASES}
\label{sec:use_cases}
This section presents three studies that draw on POLAR-Sim: object detection using a neural network (NN) trained on POLAR-Sim assets; photorealistic image synthesis followed by a sim-to-real gap assessment through the lens of a perception task; and a simulation of NASA's VIPER rover operating on POLAR-Sim virtual terrain. 

\begin{figure*}
	\centering
	\includegraphics[width=2.0\columnwidth]{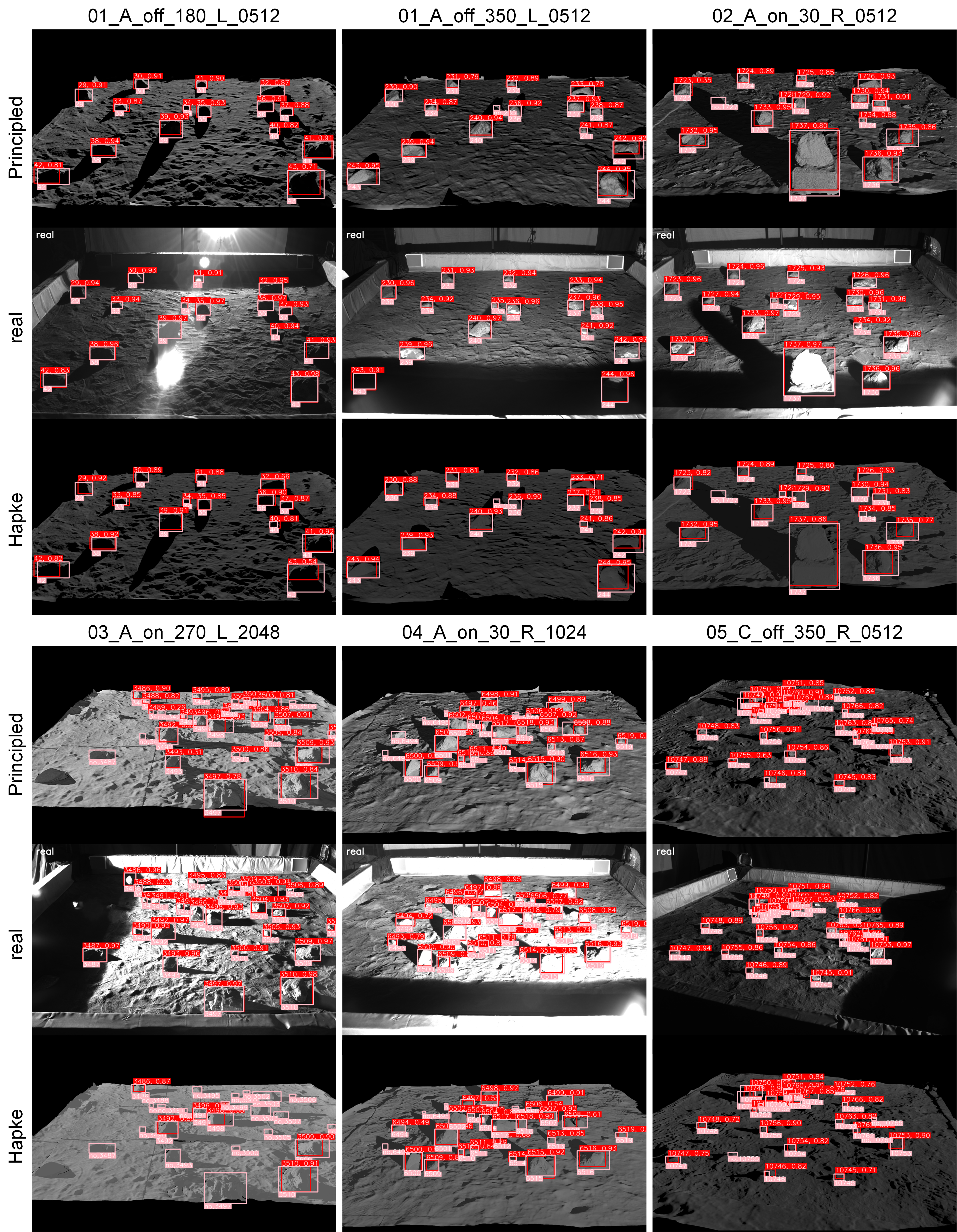}
	\caption{Illustration comparison of rock detection among real (2nd row), Principled-synthesized (1st row), and Hapke-synthesized (3rd row) images judged by YOLOv5 trained on real photos. The configuration parameters above each column are represented as: [terrain ID]\_[stereo camera position]\_[rover light status]\_[Sun azimuth]\_[(Left/Right camera]\_[exposure time]. Pink boxes are ground-truth with rock indices, and red boxes are predictions.}
	\label{fig:yolo_result1}
\end{figure*}

\begin{figure*}
	\centering
	\includegraphics[width=2.0\columnwidth]{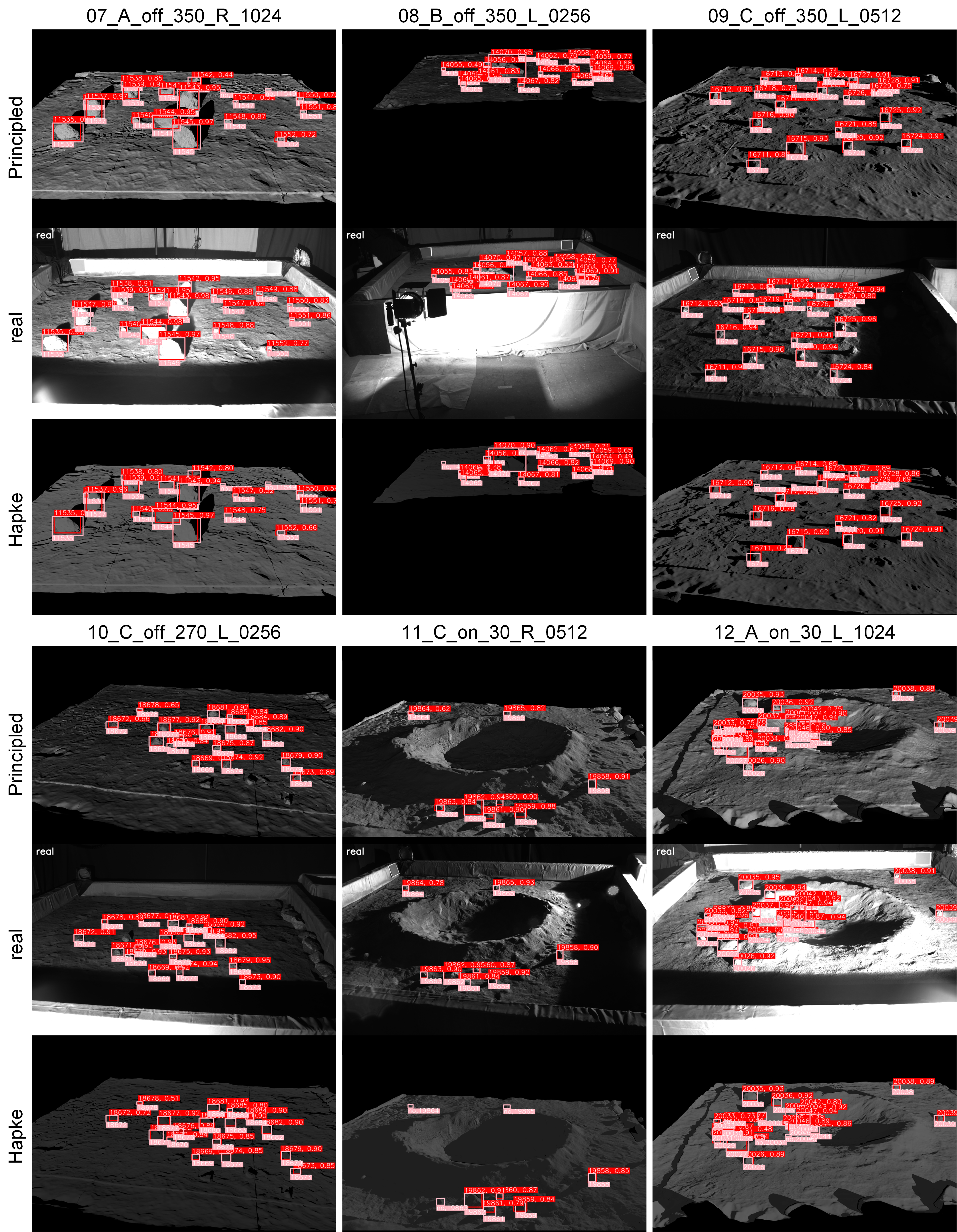}
	\caption{Illustration comparison of rock detection among real (2nd row), Principled-synthesized (1st row), and Hapke-synthesized (3rd row) images judged by YOLOv5 trained on real photos (Cont'd).}
	\label{fig:yolo_result2}
\end{figure*}

\subsection{Case I: Training an Object Detection Algorithm}
\label{subsec:train_perception}
% \SBELcomment{Bo-Hsun}. Describe how we train YOLOv5 using the labels in POLAR. Then demo the YOLOv5 solution for some other images from POLAR. Will use CPD and such to score YOLOv5. Should be max one page.

The POLAR-Sim bounding box annotations of rocks and rock shadows were used to fine-tune YOLOv5 for object detection tasks. We used 3,460 POLAR-sim annotated POLAR photos for training, and 1,730 POLAR photos for performance evaluation (note that POLAR has approximately 2,600 \textit{stereo} images). Using its pre-trained weights, YOLOv5 \cite{jocher_yolov5_2020} was fine-tuned to detect rocks and rock shadows. The model was trained and validated on the training set for 200 epochs, with the best validation weights selected. The trained YOLOv5 was then evaluated on real and synthetic photos, see Figs.~\ref{fig:yolo_result1} and \ref{fig:yolo_result2}. Since the POLAR-Sim assets allowed us to generate digital twins of POLAR photos, we trained YOLOv5 in three different ways: using POLAR photos and two different approaches to generate synthetic images (details provided in the next subsection). Table~\ref{table:yolo_result} summarizes the results, where the rock detection is reported using mean average precision (mAP) at an intersection over union (IOU) threshold of 0.5 (\textit{mAP@0.5}), mAP over several IOU thresholds from 0.5 to 0.95 (\textit{mAP@[0.5:0.95]}), and the \textit{IOU\textsubscript{mean}} value, which was obtained by choosing the predicted bounding box with the largest IOU of the ground truth bounding box label for each rock, calculating the IOU of the chosen predicted and ground truth  labels for the rock, and then averaging the value over all the rocks.

In Table~\ref{table:yolo_result}, ``T:'' indicates which dataset domain YOLOv5 was trained on: Real, Principled, or Hapke (discussed in Case II). In Case I, the focus is on ``Real,'' meaning that real POLAR photos were used for training. The ``E:'' indicates which type of photos were used to assess YOLOv5 for rock detection: Real, Principled-synthesized, or Hapke-synthesized. Note that in training and evaluation of YOLOv5 we used the digital twins of the POLAR photos. However, as discussed next, POLAR-Sim enables one to generate as many synthetic photos as needed, using different poses, exposures, camera artifact, etc.

\begin{table}[!t]
	\caption{Comparison of YOLOv5 rock detection performance results. The ``T:'' indicates how training took place; an ``E:'' indicates the evaluation set.}
	\label{table:yolo_result}
	\centering
	\resizebox{1.0\columnwidth}{!}{
	\begin{tabular}{|c|c|c|c|}
		\hline
		& \multicolumn{3}{c|}{$\uparrow$ mAP@0.5 / mAP@[0.5:0.95] / IOU\textsubscript{mean}} \\
		\hline
		& E: Real & E: Principled & E: Hapke \\
		\hline
		T: Real		& \textbf{0.987 / 0.805 / 0.849}	& 0.682 / 0.320 / 0.624				& 0.580 / 0.261 / 0.534 \\
		\hline
		T: Principled	& 0.662 / 0.318 / 0.513				& \textbf{0.893 / 0.829 / 0.893}	& 0.813 / 0.664 / 0.842 \\
		\hline
		T: Hapke		& 0.582 / 0.269 / 0.434				& 0.887 / 0.730 / 0.872				& \textbf{0.887 / 0.809 / 0.898} \\
		\hline
		\multicolumn{4}{l}{Note: the arrow pointing up $\uparrow$ in the first row means higher values are better.} \\
		\multicolumn{4}{l}{The bold number marks the best value of the row or of the column.}
	\end{tabular}
	}
\end{table}

\subsection{Case II: Lunar Image Synthesis}
\label{sub:image_synthesis}
% \SBELcomment{Bo-Hsun} Talk about Chorno::Sensor. Describe how we train YOLOv5 using the data in POLAR3D. Then demo the YOLOv5 solution for some other images from POLAR and/or POLAR3D. Should be about 0.5 pages or so.

The second use case addresses the following question: Can one synthesize images to train perception algorithms that work well on the Moon? Synthesizing photorealistic images is important because the original POLAR dataset contains only about 5,200 photos, which is a relatively small dataset. The objective is to use computer graphics techniques to generate images that closely resemble the real lunar photos. Since actual lunar photos are hardly available, the POLAR dataset serves as a proxy for what real lunar photos should look like. POLAR-Sim enables the generation of labeled lunar images on demand. With the terrain meshes available in POLAR-Sim, any camera simulator can synthesize these images. 
Herein, we used the Chrono::Sensor simulator \cite{asherSensorSimulation2021,elmquist2021modeling} to generate these lunar images. Chrono::Sensor simulates cameras to synthesize high-fidelity photorealistic images based on a bidirectional reflectance distribution function (BRDF) model and physically-based ray-tracing rendering using OptiX \cite{optixNVIDIA}. POLAR-Sim terrain meshes were imported into Chrono::Sensor, and material properties (e.g., base color, specular color, and roughness) were heuristically set by observing the appearance of the ground and rocks in the real photos. The simulated Sun, rover light, and cameras were positioned according to the POLAR dataset settings. White point lights with no spatial attenuation were used to simulate the Sun and rover light. The intensity ratio between the simulated Sun and rover light was based on the power values from the POLAR dataset document. At this point, the Chrono virtual camera could synthesize images, creating POLAR-photo synthetic counterparts using both the Principled and Hapke BRDFs.

We used the Principled and Hapke BRDFs in Chrono::Sensor to generate the synthetic images. The Principled BRDF is a simplified version of the Disney BRDF, which was proposed by Walt Disney Animation Studios and widely used in physically-based rendering \cite{burley2012physically}. The Hapke model is specifically designed to represent lunar surface reflectance characteristics \cite{hapke1981bidirectional,hapke2008bidirectional,sato2014resolved}. Details of the Hapke BRDF implementation in Chrono::Sensor are provided in \cite{batagoda2024physics-based}. The images synthesized by the Principled and Hapke BRDFs, along with real photos, were subsequently input into YOLOv5 for rock detection. Rock detection results that use the YOLOv5 trained on real photos and tested on the Principled and Hapke synthetic images are shown in Figs.~\ref{fig:yolo_result1} and \ref{fig:yolo_result2} and summarized in Table~\ref{table:yolo_result}.

In addition to using mAP to judge the synthetic dataset performance, we also used the Instance Performance Difference (IPD) metric to examine which synthetic dataset resembles the real photos better \cite{chen2024instance}. IPD quantitatively measures the domain gap between two image datasets through the lens of a perception algorithm outputs. YOLOv5 was selected as the perception algorithm, and thus trained on images synthesized by the Principled and Hapke BRDFs, and tested on all the three dataset domains (Real, Principled-synthesized, and Hapke-synthesized). As also demonstrated here, which photo dataset is used to train YOLOv5 does affect its performance \cite{liu2020neural}. The training setups were the same as explained in Sec.~\ref{subsec:train_perception}, except that the ground truth bounding boxes of rocks were automatically generated in the synthetic images. The mAP results of YOLOv5, trained and tested on the three different dataset domains, are listed in Table~\ref{table:yolo_result}. The IPD values between the real photos and synthetic images from the two BRDFs are listed in Table~\ref{table:IPD_result}.

\begin{table}[!t]
	\caption{Comparison of IPD results.}
	\label{table:IPD_result}
	\centering
	\resizebox{1.0\columnwidth}{!}{
		\begin{tabular}{|c|c|c|c|}
			\hline
			& \multicolumn{3}{c|}{$\downarrow$ Instance Performance Difference (IPD)} \\
			\hline
			& E: $\|\text{Principled - Hapke}\|$ & E: $\|\text{Real - Hapke}\|$ & E: $\|\text{Real - Principled}\|$ \\
			\hline
			T: Real 		& - 				& 0.3152	& \textbf{0.2256}	\\
			\hline
			T: Principled	& \textbf{0.0511} 	& - 		& 0.3808			\\
			\hline
			T: Hapke		& \textbf{0.0261} 	& 0.4638	& - 				\\
			\hline
			\multicolumn{4}{l}{Note: the arrow pointing down $\downarrow$ in the first row means lower values are better.} \\
			\multicolumn{4}{l}{The bold number marks the best value of the row.}
		\end{tabular}
	}
\end{table}

\subsection{Case III: Rover Simulations in POLAR-Sim Virtual-Worlds}
\label{subsec:rover_simulation}
The last technology demonstration highlights how the POLAR-Sim assets facilitate model-based synthesis of a rover's autonomy stack. In addition to POLAR-Sim, this case study uses a Chrono digital twin of the VIPER. Effectively, the rover drives over the POLAR-Sim mesh assets, where the latter are used to define a deformable terrain profile. The terrain dynamically deforms as the rover traverses it. The simulation integrates the POLAR-Sim meshes, the camera simulator, the Chrono dynamics simulation engine, and the YOLOv5 NN of Case I.

The terrain and rock meshes of three POLAR scenarios -- 4, 1, and 11, in that order -- were combined into one patch of ``virtual world,'' where the simulated VIPER was operated using Chrono::Vehicle \cite{chronoVehicle2019}, moving from the east to west sides of the patch. A point light was added to simulate the Sun. The ground was modeled as deformable through the Soil Contact Model (SCM) to capture the terramechanics interactions between the rover and the terrain \cite{chronoSCM2019,chronoSCM2022}. A proportional controller (P-control) was employed for steering the VIPER to run along the middle line of the terrain.

During simulation, virtual cameras registered images. Four cameras were positioned next to the four wheels of the rover, respectively, to closely observe the terramechanics between the wheel and terrain. Two other fixed third-person-view cameras monitored the rover movement from the left and right sides, respectively. A virtual camera was mounted at the front-end of the rover for hazard detection using the similar camera pose in the POLAR dataset, thus feeding the synthetic images into YOLOv5 for rock and shadow recognition. The Deep Star Map HDR image was used as the environment map \cite{nasa2020deep}. Four different Sun's directions (East, Southeast, Southwest, and West) were tested to observe how lighting and shadow changes impact YOLO's performance. Additionally, in Chrono::Sensor, both the Principled and Hapke BRDFs were respectively used to observe rendering differences.

\begin{figure*}[!t]
	\centering
	\includegraphics[width=2.0\columnwidth]{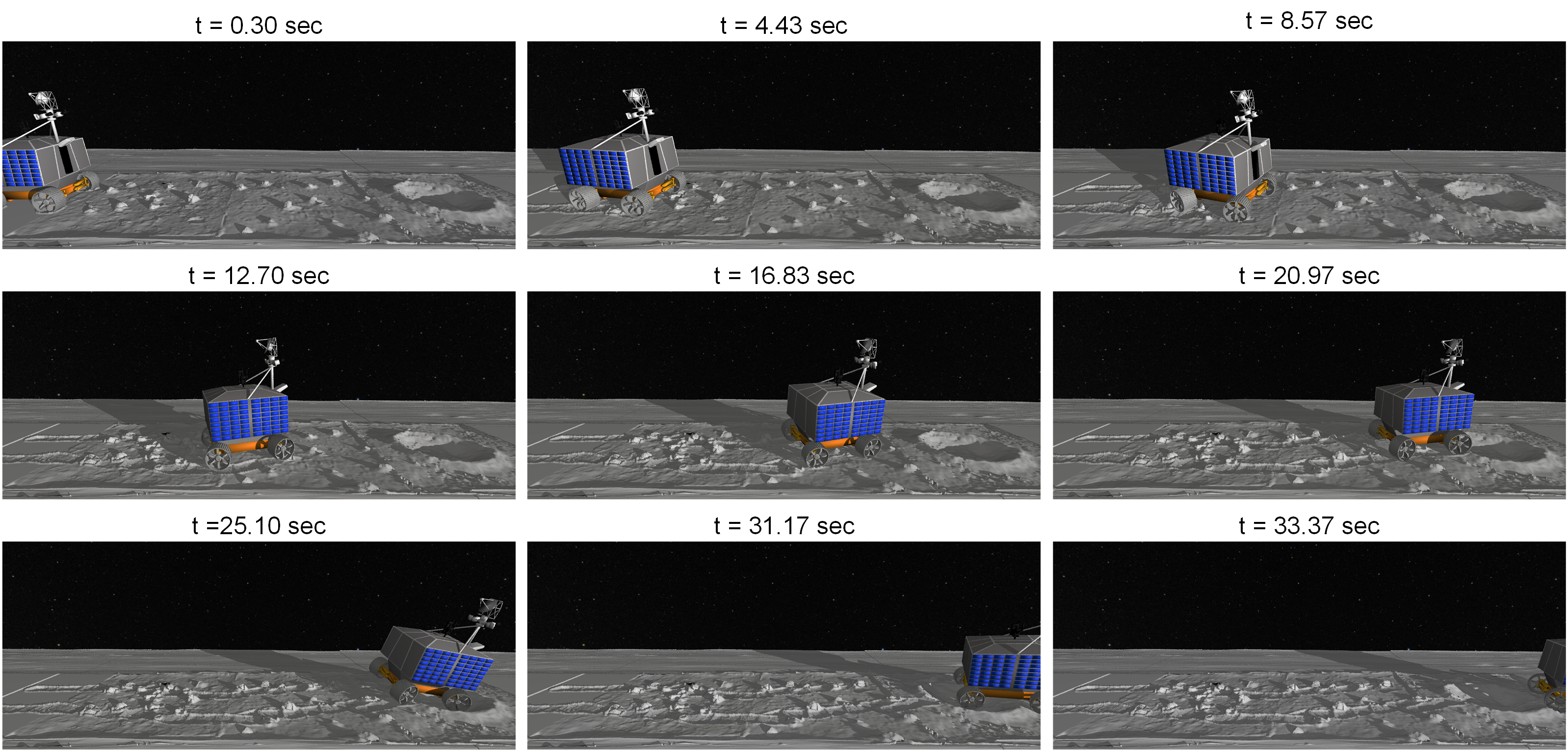}
	\caption{A sequence of images from the left third-person-view camera showing the rover's movement. Images are ordered from left to right, top to bottom, and are horizontally flipped for visualization here.}
	\label{fig:img_series}
\end{figure*}

The simulation setup enables integrated testing of perception, trajectory planning, and control algorithms all in a dynamic-model-based and data-driven simulation framework. With the POLAR-Sim digital assets, one can have an all-in-one framework supporting sensor, vehicle dynamics, and terramechanics simulation. The uploaded media files demonstrate the VIPER simulation traversing a section of deformable lunar terrain, populated with rocks and a big crater. Observations reveal that when encountering rocks too large to run over, the rover's wheels just slipped around and circumvented the obstacles. In this way, the videos show that the VIPER successfully navigated across the three terrain patches and ran over all collidable rocks and one large crater. A sequence of images from the left third-person-view camera is presented in Fig.~\ref{fig:img_series} for process illustration.

\section{DISCUSSION}
\label{sec:result_discussion}
\subsection{Discussion of Cases I and II}
\label{subsec:image_discussion}
% \SBELcomment{Bo-Hsun}: Highlight the accomplishments we report in this contribution.

\begin{figure}[!t]
	\centering
	\includegraphics[width=1.0\columnwidth]{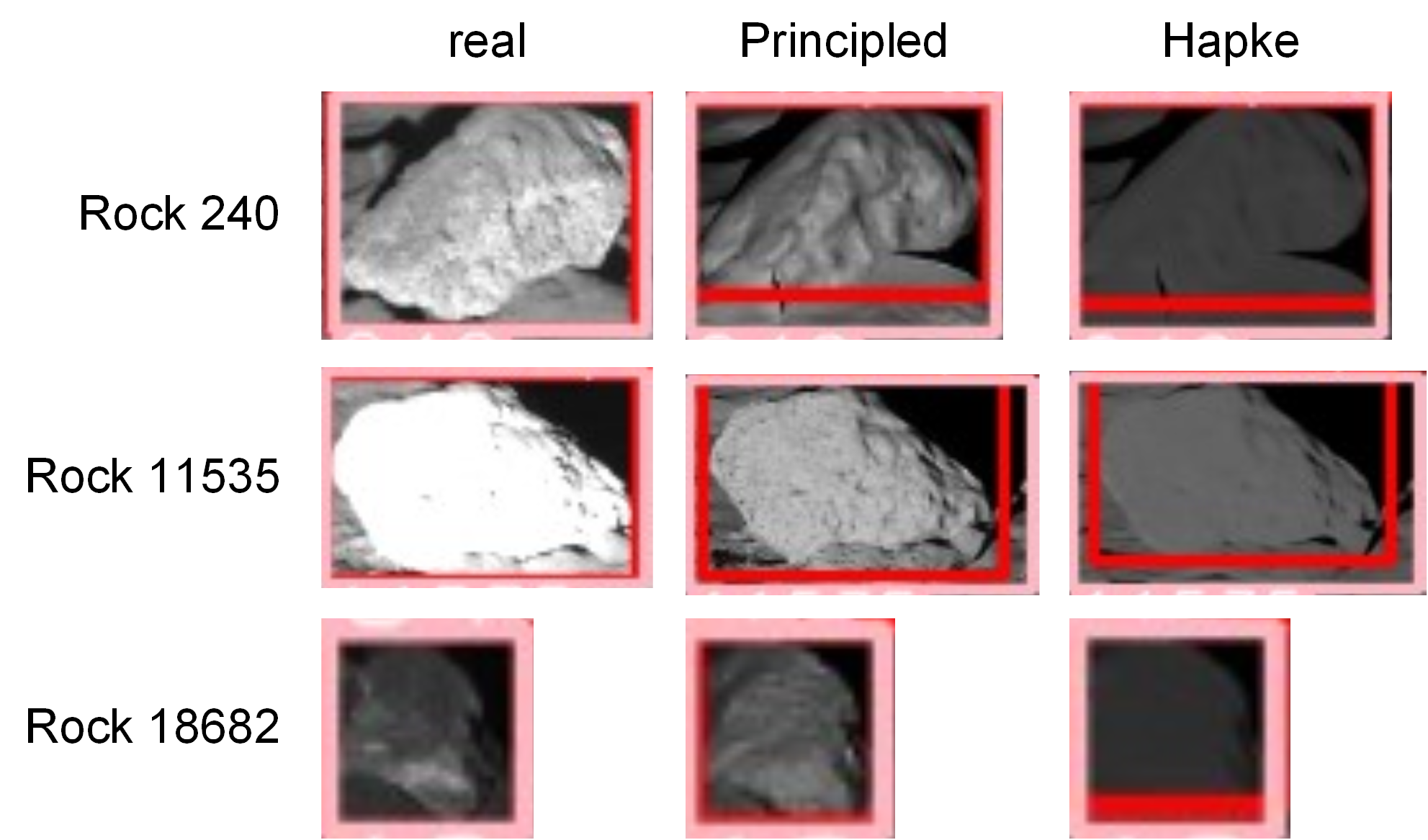}
	\caption{Rocks under the opposition effect.}
	\label{fig:oppo_effect}
\end{figure}

Leveraging the POLAR-Sim labeled bounding boxes and its digital twins of the POLAR terrain scenarios, we synthesized images resembling the real photos of the POLAR dataset. Subsequently, we performed rock detection to assess, in a statistical sense, how YOLOv5 performs on real photos and synthetic images. Note that there is nothing special about choosing YOLOv5 -- its selection was due to its wide user base and known good performance. Indeed, a different user might select to compare the synthetic vs. real pictures through the lens of a different ``judge,'' e.g., the IGEV algorithm \cite{xu2023iterative} for depth estimation in stereo images. Figures~\ref{fig:yolo_result1} and \ref{fig:yolo_result2} summarize the test results. Qualitatively, both the Principled (which draws on \cite{burley2012physically}) and Hapke BRDF-synthesized images look closely similar to the real photos. The synthetic images successfully capture the light and shadow structures of the real photos and have similar silhouettes of rocks, pits, hills, and terrain undulation, compared to their real counterparts. The Principled BRDF renders rocks textures more expressively, while the Hapke model loses texture on the rock surfaces when the Sun and camera are pointing in the same direction, a phenomenon known as the \textit{opposition effect} and shown with Rocks 240 in Terrain 1, 11535 in Terrain 7, and 18682 in Terrain 10 from Figs.~\ref{fig:yolo_result1} and \ref{fig:yolo_result2} (see Fig.~\ref{fig:oppo_effect} for zoomed-in visualization). Unsurprisingly, the Principled BRDF produces results that appear more similar to the real photos, which were taken on Earth, than the Hapke BRDF. Note that the YOLOv5 trained on real photos has high detection accuracy in conjunction with the synthetic images, which yet again speaks to the accuracy of the terrain digitization and photorealistic image synthesis in this work.

A quantitative comparison is shown in Tables~\ref{table:yolo_result} and \ref{table:IPD_result}. Unsurprisingly, when YOLOv5 is trained using photos from a particular domain, say real photos, it will yield the best detection performance when presented with photos from the same dataset. Beyond this basic observation, when crossing domains, YOLOv5 trained on real photos did better on detecting rocks in Principled-synthesized photos than in Hapke ones, indicating that as far as YOLOv5 is concerned, the Principled-synthesized rocks are more similar to the real rocks. This is further supported by the fact that the IPD between the real and Principled-synthesized rocks is 28.43\% less than the IPD between the real and Hapke-synthesized rocks when judged by the YOLOv5 trained on real photos. Besides, considering the purpose of using synthetic images to train perception algorithms for real-world visual detection, YOLOv5 trained on the Principled-synthesized images detected 13.74\%, 18.21\%, and 18.20\% more rocks than when trained on the Hapke-synthesized images, based on the mAP@0.5, mAP@[0.5:0.95], and IOU\textsubscript{mean} metrics, respectively. This confirms again that the Principled BRDF better fits the real photos in the POLAR dataset than the Hapke BRDF. We posit that this discrepancy may stem from the regolith simulant used in the POLAR terrain, which lacks the fine and porous structure and optical reflective properties of the real lunar sand; i.e., the POLAR photos were taken on Earth using material that does not have the geometry and morphology (and thus light reflective properties) of actual lunar regolith. Moreover, the POLAR terrain's atmospheric conditions are Earth-like, where ambient light scattered from the air exists, unlike the almost-vacuum state on the Moon. 

Finally, note that in the last two rows in Table~\ref{table:IPD_result}, the IPD between rocks synthesized by the Principled and Hapke BRDFs is much smaller than the IPD between real and synthetic rocks, when judged by the YOLOv5 trained on images synthesized by either BRDF. This shows that the Principled and Hapke synthetic domains are relatively very close to each other, while both still displaying a slight gap relative to the real domain. We posit that this is an artifact tied to the observation that the ray-tracing rendering system in Chrono::Sensor does not perfectly model the real-world reflectance details, despite appearing qualitatively similar. The synthetic images exhibit less contrast compared to the real photos, and the BRDF's limited representation of the self-shadowing and self-masking effects on the small surface facets results in less vivid rock patterns. In some cases, the synthetic images also miss artifacts seen in the real photos, such as very bright specular highlights and lens flare.

\subsection{Demonstration Result of Case III}
\label{subsec:demo_result_discuss}
\begin{figure}[!t]
	\centering
	\includegraphics[width=1.0\columnwidth]{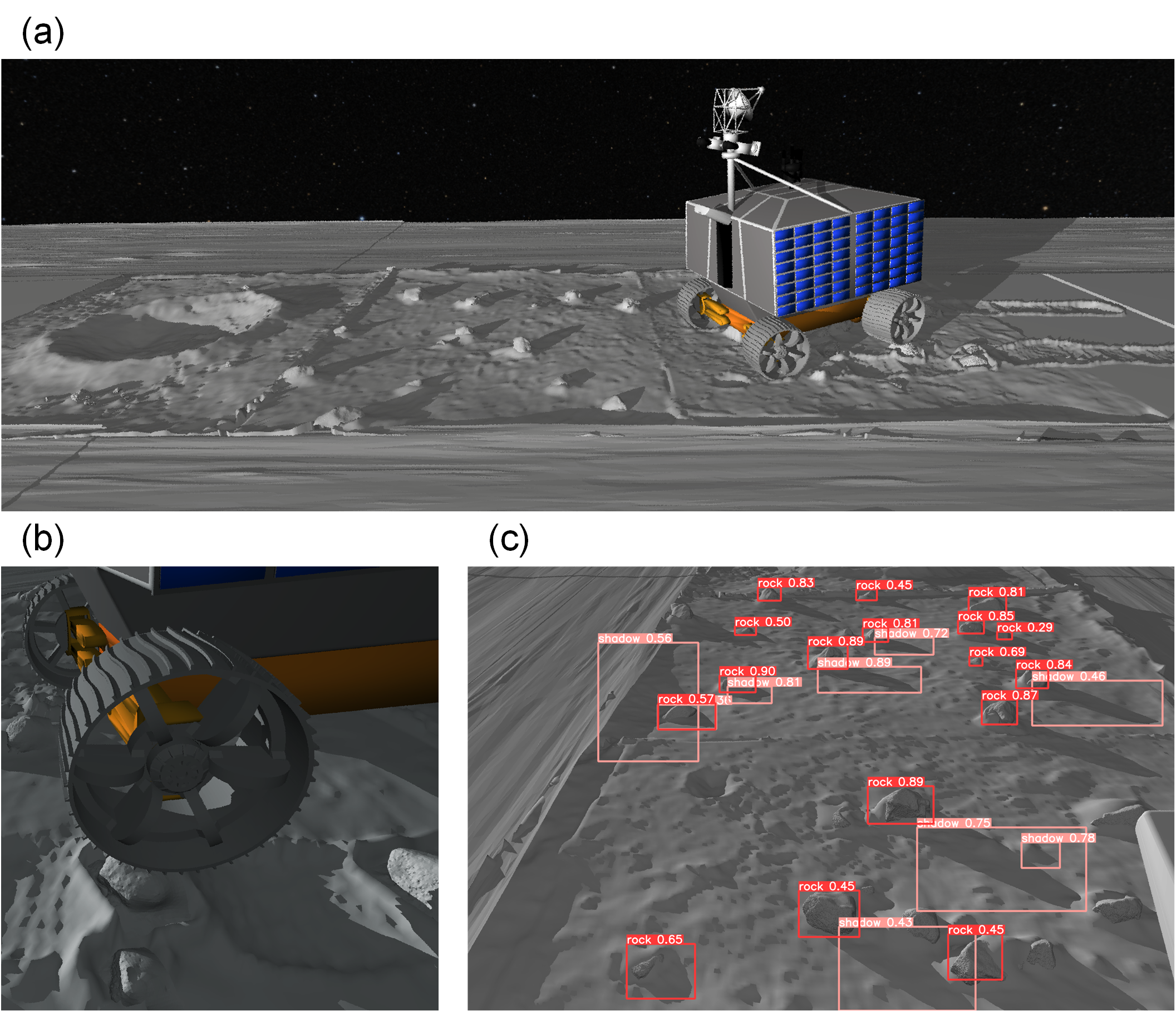}
	\caption{Illustrations of VIPER traversing digitized terrain of the POLAR dataset from right to left. (a) Third-person-view from the left side, (b) right-back-wheel-attached camera, and (c) front-end camera recognizing rocks and shadows by YOLOv5 for hazard detection.}
	\label{fig:cover_image}
\end{figure}

\begin{figure}[!t]
	\centering
	\includegraphics[width=1.0\columnwidth]{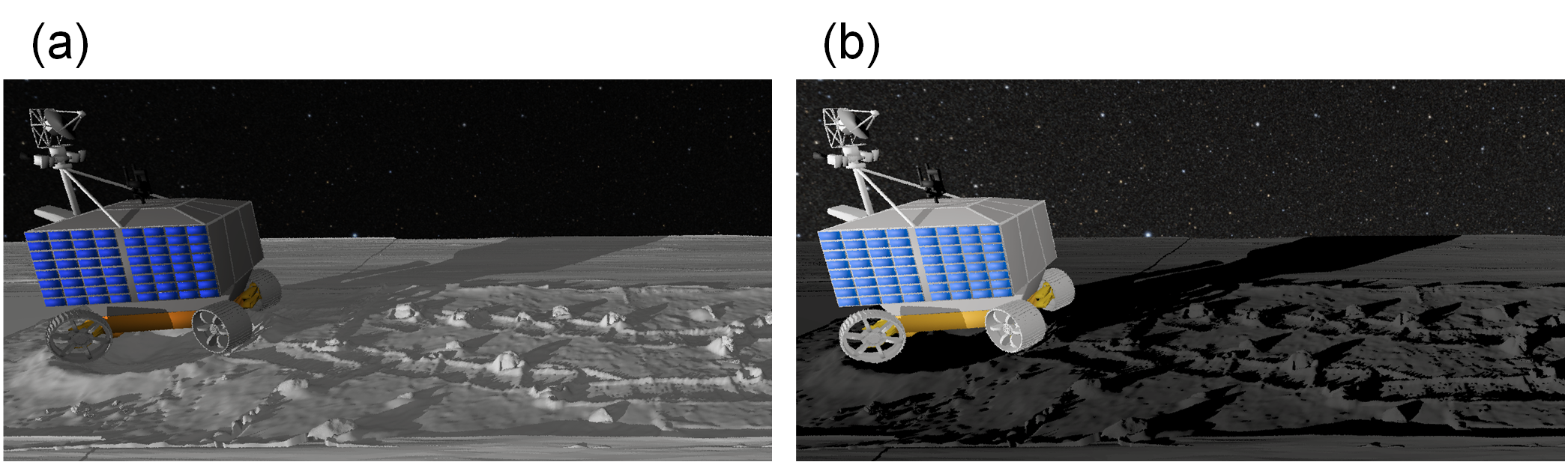}
	\caption{Different rendering modes of the (a) Principled and (b) Hapke BRDFs in dynamic simulations.}
	\label{fig:BRDF_dynamic_compare}
\end{figure}

In this test, the virtual rover moves around in a digital twin stitched together using POLAR terrain scenarios while using a virtual camera to take synthetic images that change according to the locomotion of the VIPER. Images from the virtual cameras are displayed in Fig.~\ref{fig:cover_image}. Figure~\ref{fig:cover_image}(a) presents a third-person-view camera recording the vehicle's locomotion. Figure~\ref{fig:cover_image}(b) shows the right-back-wheel-attached camera when the VIPER runs over a rock in POLAR-Sim, capturing the high-fidelity terramechanics interaction between the wheel and the rock. Supported by the SCM deformable terrain model, which uses a Bekker-Wong type terramechanics model~\cite{chronoSCM_JCND_2023}, this demo highlights the potential of registering how different soil parameters or wheel topologies influence the motion of the rover. Lastly, Fig.~\ref{fig:cover_image}(c) shows the front-end camera view with the YOLOv5 detection results, demonstrating the vehicle's ability to register rocks and shadows while traversing the virtual terrain. These images showcase the feasibility of developing and testing data-driven visual hazard detection algorithms alongside vehicle dynamics and terramechanics simulations in the open-source Chrono simulator \cite{chronoOverview2016,chronoProjectsWebSite}. Finally, Fig.~\ref{fig:BRDF_dynamic_compare} illustrates the differences between the Hapke and Principled BRDFs during the dynamic simulation. The Hapke BRDF, designed specifically for lunar surface reflectance, rendered darker shadows and more pronounced back-scattering effects on the terrain, compared to the Principled BRDF.

The complete collection of simulation videos is available at \cite{chen2024polarsim_videos}. All videos were made at 0.5x speed. Videos are named in the following format: [camera viewpoint]\_[Sun ID]\_[BRDF]\_ [exposure time].mp4. The [camera viewpoint] includes two third-person-view cameras from the left and right sides of the VIPER, \textit{CamBirdViewLeft} and \textit{CamBirdViewRight}, four wheel cameras, \textit{WheelCam\_RightFront}, \textit{WheelCam\_RightBack}, \textit{WheelCam\_LeftFront}, and \textit{WheelCam\_LeftBack}, and the front-end camera, \textit{front\_end\_cam}, respectively. The [Sun ID] annotates the Sun's directions, where 1, 2, 3, and 4 represent East, Southeast, Southwest, and West, respectively. \textit{hapke} or \textit{default} in the [BRDF] specifies either the Hapke or Principled BRDF in Chrono::Sensor. And [exposure time] is set to 0256, 0512, or 1024 milliseconds.

\subsection{Similar Work, and Relation to the State of the Art}
\label{subsec:related_work}
%\SBELcomment{Nevindu} to take a first shot. No more than 0.5 pages. Short subsection. Just look around and mention what other data sets are available out there. We should look beyond extraterrestrial datasets, and include Cityscapes \cite{cordts2016cityscapes} and the like. Test reference~\cite{asherSensorSimulation2021}.

In extraterrestrial exploration research, several simulators have integrated robot dynamics simulation with photorealistic virtual camera rendering \cite{pangu2004, brochard2018scientific, muller2021photorealistic}, which is critical for training both human operators and data-driven visual algorithms under challenging light conditions. For example, NASA's planetary rover simulator, based on Gazebo \cite{koenig2004design}, supports rover navigation, system monitoring, and scientific instrument simulations \cite{allan2019planetary}. It utilizes the Ogre3D engine to simulate a virtual camera and incorporates a modified Hapke BRDF \cite{sato2014resolved} to model lunar optical behavior. Similarly, the \textit{Unreal Robot Simulation (URSim)}, developed by the German Aerospace Center (DLR), uses Unreal Engine 4 to render photorealistic Martian and lunar landscapes in real time for planetary robotic navigation \cite{sewtz2022ursim}. Another physics-based simulator, JPL's \textit{DARTS ROAMS}, focuses on accurate vehicle dynamics and sensor simulations for planetary rovers \cite{Roams2001}. These simulators focus on replicating extraterrestrial visual and dynamic environments, like Chrono \cite{chronoOverview2016}, but this is the first effort in which the synthetic images have been compared to the POLAR real photos in the context of a validation effort.

Several well-known labeled datasets, such as \cite{neuhold2017mapillary, huang2018apolloscape, bdd100k}, feature real-world photos to benchmark visual perception algorithms. Notably, the \textit{Cityscapes} and \textit{KITTI} datasets are widely used in autonomous vehicle development \cite{cordts2016cityscapes, geiger2013vision}. These datasets include photos captured by cameras mounted on cars and depict urban environments with pedestrians, vehicles, and buildings. However, they lack scenarios representing the more challenging off-road or wilderness conditions relevant to space exploration tasks.

Several datasets tailored to extraterrestrial environments have been curated and are available in the literature. For instance, the dataset from Mt. Etna in Sicily is considered a good analogue to lunar conditions in terms of soil properties and appearance \cite{vayugundla2018datasets}. The \textit{Artificial Lunar Landscape Dataset} features 9,766 computer-synthesized images of rocky lunar landscapes \cite{lunarLandscapeDataset}. \textit{Deep Mars} provides a searchable database of over 22 million actual Martian landscape images \cite{wagstaff2018deep}, and the MADMAX dataset, similar to POLAR, includes sensor data recorded during experiments in the Moroccan desert \cite{meyer2021madmax}. The POLAR dataset has also been utilized in various works, such as creating the simulated scenarios to develop a shared mixed-reality system for human-robot collaboration in lunar exploration \cite{ji2024on-site}, and initiating the \textit{Disparity Computation Framework} that accommodates different stereo camera algorithms using a single standard \cite{da_silva_vieira2019disparity, vieira2021three-layer}. POLAR-Sim distinguishes itself in two key aspects: (\textit{i}) it provides labeled assets for photos specifically crafted to replicate lunar environments, which enables the development and validation of data-driven perception algorithms; and (\textit{ii}) it facilitates synthesizing any number of custom lunar synthetic photos when combined with a camera simulator such as Chrono::Sensor.

%Finally, note that existing digital elevation models (DEMs) of the lunar surface (which were scanned, for instance, by the Lunar Orbiter Laser Altimeter (LOLA) \cite{BARKER2016346}) provide detailed geometry, they are not accompanied by real photos as ground truth for rendering validation. Thus, although in a typical vehicle simulation for lunar exploration one can leverage the DEM models in combination with the Hapke BRDF for virtual camera rendering, this lacks the ground truth information necessary to evaluate the quality of the rendered images. In contrast, the POLAR dataset includes real photos for validation. Additionally, the POLAR dataset specifies the number of rocks in each terrain scenario, aiding annotators in accurately locating and separating rocks from the ground. When treating the rocks as obstacles, such rock separation is crucial for training data-driven hazard detection algorithms and for constructing simulation scenarios with assigning distinct visual and dynamic characteristics for meshes of rocks and ground. Thus, our POLAR-Sim dataset is better suited for integrating camera and vehicle dynamics simulations compared to using the actual lunar DEM models alone.

\section{CONCLUSION}
\label{sec:conclusion}
This contribution introduces \textit{POLAR-Sim}, a database of labels and digital assets that augments NASA's POLAR dataset of stereo photos. POLAR-Sim includes: (\textit{i}) manually labeled bounding boxes and semantic segmentation maps of the ground, rocks, and rock shadows for approximately 23,000 rocks across approximately 2,600 pairs of stereo POLAR dataset photos; and (\textit{ii}) manually separated ground and rock meshes, providing digital twins for all POLAR terrain scenarios. We demonstrated the use of POLAR-Sim in three instance: ($a$) to train the YOLOv5 object detection neural network using the newly generated label annotations; ($b$) to synthesize, using Chrono::Sensor's Principled and Hapke BRDFs, the digital twins of the POLAR images; and ($c$) to simulate the VIPER locomoting in a virtual world stitched together from digital twins of selected POLAR terrain scenarios. 

Future improvements include automating the generation of ground-truth shadow labels in synthetic images. Additional fine-tuning of the database is necessary to make the ground and rock meshes more consistent. Currently, some mesh poses exhibit slight biases relative to their real-world counterparts, leading to object positional discrepancies between the real and synthetic pictures. Finally, \textit{automatically} determining the material parameters and texture maps for the ground and rocks in the Principled and Hapke BRDF models is the subject of an ongoing research effort.

%\bibsection*{REFERENCES}
\bibliographystyle{ieeetr}
\bibliography{BibFiles/refsSensors,BibFiles/refsML-AI,BibFiles/refsAutonomousVehicles,BibFiles/refsChronoSpecific,BibFiles/refsSBELspecific,BibFiles/refsMBS,BibFiles/refsCompSci,BibFiles/refsTerramech,BibFiles/refsFSI,BibFiles/refsRobotics,BibFiles/refsDEM,BibFiles/refsGraphics}

\end{document}